\documentclass[letterpaper]{article} 
\usepackage[preprint]{aaai2027} 
\usepackage[hyphens]{url} 
\usepackage{graphicx} 
\usepackage{natbib} 
\usepackage{caption} 
\usepackage{amsmath}
\usepackage{amssymb}
\usepackage{booktabs}
\usepackage{colortbl}
\usepackage{arydshln}
\usepackage{xspace}

\newcommand{\method}{GAP-SAM\xspace}
\newcommand{\dataset}{COCO-ControlNet\xspace}

\title{GAP-SAM: A Global Artifact Prior for Generalizable AI-Generated Image Manipulation Localization}

\author{
Haozhen Yan\textsuperscript{\rm 1}\equalcontrib,
Siyuan Shan\textsuperscript{\rm 1}\equalcontrib,
Zijian Yu\textsuperscript{\rm 2},
Youqi Wang\textsuperscript{\rm 3},
Yan Hong\textsuperscript{\rm 2}, \\
Jun Lan\textsuperscript{\rm 2}\thanks{Corresponding authors.},
Jianfu Zhang\textsuperscript{\rm 1}\footnotemark[2]
}

\affiliations{
\textsuperscript{\rm 1}Shanghai Jiao Tong University,
\textsuperscript{\rm 2}Ant Group, 
\textsuperscript{\rm 3}Shenzhen University,\\
\small \{orion810, c.sis\}@sjtu.edu.cn, lanjun\_yelan@163.com
}

\begin{document}

\maketitle

\begin{abstract}

AI-generated image manipulation localization identifies edited pixels, but its OOD performance lags behind image-level detection partly because pixel supervision entangles forensic evidence with dataset-specific mask geometry and semantic boundaries.
Extending image-level distribution alignment to localization, we construct COCO-ControlNet with source-image Canny edges and depth maps to align semantics and geometry, improving OOD performance across multiple localizers.
Yet tighter Mask-VAE Reconstruction Alignment (Mask-VAE) underperforms COCO-ControlNet, showing that VAE reconstruction artifacts transfer poorly to local diffusion-inpainting artifacts.
We also identify \emph{boundary adhesion}, where fine-tuned segmentation models snap predictions to semantic object contours rather than true manipulation boundaries.
These findings motivate GAP-SAM, which encodes an image and its frozen VAE reconstruction into a global artifact token and injects it into SAM3's feature pyramid via zero-gated FiLM before pixel decoding.
Without prescribing a spatial region, this token modulates dense decoding to preserve localization while suppressing semantic-boundary shortcuts.
Across six datasets, GAP-SAM averages 79.8 Pixel-F1, outperforming the strongest prior method by 12.6 points.
It also performs best at every tested severity of JPEG compression, Gaussian blur, and resizing.

\end{abstract}

\section{Introduction}

Rapid advances in generative models have made sophisticated image manipulation accessible through simple user instructions, greatly reducing the required cost and expertise~\cite{sun2024rethinkingediting,wang2025opensdi}.
This growing accessibility raises concerns about data integrity and information security~\cite{sun2024rethinkingediting,guan2019mfc}.
Image Manipulation Localization (IML) identifies manipulated regions at the pixel level to expose deceptive content.
However, existing methods often degrade substantially in real-world scenarios~\cite{Zhu2026RITACVPRF,Qu2024ModernIMLCVPR,yan2025cocoinpaint,giakoumoglou2025diquid}, making robust OOD generalization a central challenge.

Prior work attributes this problem to training-data distribution shifts, but how to mitigate them through dataset and model design remains open.
Our review of image-level detection (Sec.~\ref{sec:image_detection_alignment}) shows that \textbf{tighter alignment between real and synthetic distributions reduces dataset-specific biases and promotes more transferable forensic evidence}.
Does pixel-level localization likewise benefit from distribution alignment, and might it require even tighter alignment?
Compared with image-level classification, pixel-level localization is more prone to overfitting non-causal factors because it must assign forensic evidence to individual pixels.
Shared-backbone multitask experiments support this hypothesis: localization suffers a larger cross-domain performance drop than image-level prediction.

Prior datasets typically apply masked edits to semantic objects in real images, introducing non-causal factors such as semantic boundaries.
Manipulation models also generate edited regions using only the prompt and unmasked context; without geometric constraints, generation often fails.
To improve alignment, we construct \dataset using Canny edges and depth maps from source images as additional generation conditions.
All seven localizers retrained on \dataset achieve higher mean OOD Pixel-F1 than their official checkpoints, demonstrating the effectiveness of \dataset as a training corpus for OOD localization.

This raises two questions: \textbf{does tighter alignment always improve pixel-level localization, and is there a better alignment strategy?}
We introduce an extreme alignment strategy, Mask-VAE, that replaces the original masked region with its VAE-reconstructed counterpart to create a locally reconstructed fake sample.
However, Mask-VAE underperforms \dataset, showing that maximizing pixel-level alignment does not guarantee optimal generalization.
We attribute this limited transferability to two mismatches in spatial support and artifact formation.
Spatially, VAE reconstruction affects the entire image, whereas Mask-VAE restricts global decoder traces to a mask without adapting them to its boundary or context.
By contrast, genuine local diffusion inpainting conditions iterative denoising on the unmasked pixels and other signals.
Its artifacts therefore depend on the denoising trajectory, context, noise schedule, mask boundary, and prompt, none of which a simple VAE encode--decode operation reproduces.

Although VAE reconstruction is unsuitable as dense spatial supervision, it can still guide local predictions as a global prior, particularly for non-semantic manipulated regions.
Directly fine-tuned segmentation models often find the approximate manipulated region but snap their predictions to the semantic contours of generated objects or anomalous regions, overlooking the true manipulation boundary.
We refer to this failure mode as \emph{boundary adhesion}.
We attribute it to a mismatch between semantic object boundaries learned during segmentation pretraining and true manipulation boundaries.

Exploiting segmentation models' spatial representations while suppressing their semantic-boundary shortcuts offers a promising path to OOD generalization.
To this end, we propose \method, which fuses the input image with its reconstruction from a frozen VAE to form a global artifact prior.
By guiding the mask decoder toward forensically supported boundaries rather than those learned solely from training data, ground-truth masks, and localization loss, this prior reduces reliance on dataset-specific correlations.

Specifically, \method injects this prior into the SAM3 feature pyramid~\cite{carion2025sam3} via zero-gated FiLM before pixel decoding, while an auxiliary image-level objective strengthens forensic evidence without spatial supervision.
We evaluate on held-out generators and mask types and under common post-processing operations.
\dataset consistently improves the OOD performance of existing localizers, whereas Mask-VAE supervision remains less effective despite tighter pixel pairing.
Across six datasets, \method achieves an average Pixel-F1 of 79.8, outperforming the strongest prior method by 12.6 points.
It also achieves the best Pixel-F1 at every tested severity of JPEG compression, Gaussian blur, and resizing.

Our contributions are threefold:
\begin{itemize}
    \item We show that manipulation localization also requires distribution alignment.
    A matched comparison with Mask-VAE shows that closer pixel pairing does not necessarily improve generalization and may instead provide weaker spatial supervision.
    \item We construct \dataset, which uses Canny edges and depth maps from source images as generation conditions to improve semantic and geometric alignment.
    \item We propose \method, which injects VAE reconstruction evidence into the feature pyramid as a global prior before pixel decoding to mitigate boundary adhesion and improve OOD generalization.
\end{itemize}

\section{Related Work}

\subsection{Image Manipulation Localization}

Conventional localizers model manipulation traces, spatial dependencies, compression artifacts, noise and boundary cues, and spatial-channel correlations~\cite{wu2019mantranet,Hu2020SPANECCV,Hao2021TransForensicsICCV,Kwon2021CATNetWACV,chen2021mvss,liu2022pscc}.
Subsequent work incorporates object context~\cite{Wang2022ObjectFormerCVPR}, forensic fingerprints~\cite{guillaro2023trufor}, multi-view representations~\cite{Li2024UnionFormerCVPR}, diffusion priors~\cite{Yu2024DiffForensicsCVPR}, decision learning~\cite{Peng2024CoDETIFS}, universal segmentation~\cite{kwon2024safire}, sparse non-semantic representations~\cite{su2025can}, and multiscale discrepancies~\cite{zhu2025mesoscopic,Chen2026FRDNetAAAI}.
Recent work targets large-scale generative and realistic edits~\cite{Qu2024ModernIMLCVPR,Zhu2026RITACVPRF}, while benchmarks reveal substantial cross-model degradation~\cite{chen2025gim,wang2025opensdi,yan2025cocoinpaint}; COCO-Inpaint also shows sensitivity to mask shape and area, suggesting mask geometry as a shortcut and motivating decoupling from semantic objects.

Foundation-model work adapts SAM with perturbation-derived forensic cues~\cite{Lek2026DetectiveSAMICLR} and investigates multimodal models for localization~\cite{Huang2025MMIMLICIP,Guo2026RethinkingVLMsCVPRF}; SIDA and FakeShield, for example, project LISA-style final-layer \texttt{<SEG>} representations into a SAM mask decoder~\cite{Huang2025SIDA,xu2025fakeshield,Lai2024LISACVPR}.
This cross-space interface may lose forensic cues and favor semantic objects, limiting localization of background regions and arbitrarily shaped manipulations.

\subsection{Distribution Alignment for AI-Generated Image Detection}
\label{sec:image_detection_alignment}
OOD generalization is particularly challenging for AI-generated image manipulation localization. 
From a complementary perspective, image-level detection reveals a consistent trend: \textbf{the closer generated images align with the real-image distribution, the less detectors rely on shortcut cues and the stronger their OOD robustness becomes}.
Grommelt et al.~\cite{grommelt2024fake} show that detectors may exploit non-causal cues from file formats, resolution, compression, and acquisition pipelines rather than transferable generation artifacts.
More broadly, a line of work reveals a clear progression toward closer alignment between generated and real image 1distributions.
GenImage~\cite{zhu2023genimage} introduces \textbf{class alignment} by pairing real and generated images within the same ImageNet categories. SemGIR~\cite{yu2024semgir} advances to \textbf{prompt alignment} by synthesizing images from real image captions. B-Free~\cite{guillaro2024bias} adopts \textbf{inpainting alignment} to reduce content bias. DIRE~\cite{wang2023dire} and DRCT~\cite{chen2024drct} further use \textbf{diffusion reconstruction} to improve sample alignment. AlignedForensics~\cite{rajan2025aligned}, REM~\cite{liu2025beyond}, and M2EA~\cite{liu2026m} simplify this paradigm through \textbf{VAE reconstruction}, which constructs synthetic counterparts aligned with real images.
Beyond pixel-space alignment, DDA~\cite{chen2025dual} observes VAE alignment may introduce additional mismatch in the frequency domain, and therefore proposes \textbf{dual alignment} in both pixel and frequency spaces to suppress high-frequency bias.
Overall, tighter distribution alignment strengthens OOD generalization in image-level detection, suggesting the same principle may benefit manipulation localization.

\section{Motivation and Analysis}

\subsection{Why Localization Generalizes Worse}

\begin{figure}[t]
\centering
\includegraphics[width=\columnwidth]{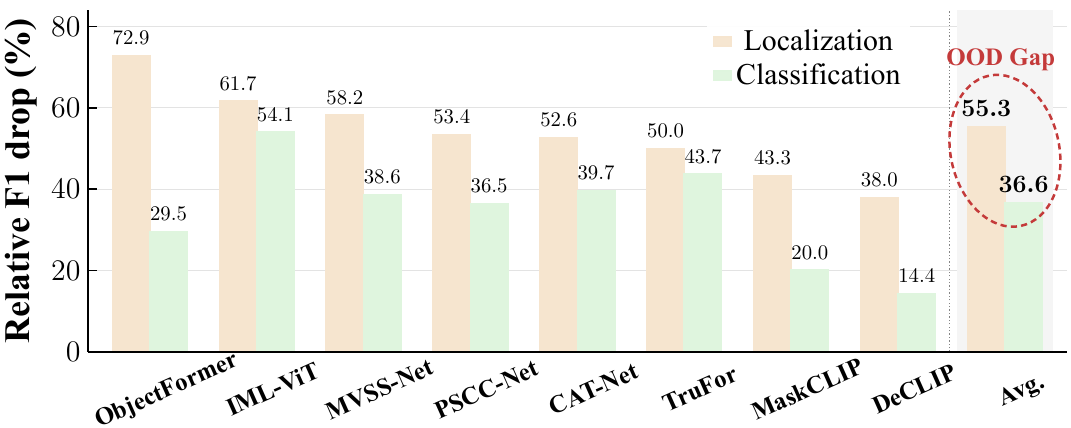}
\caption{Relative IID-to-OOD F1 drops on OpenSDI; localization drops more than classification, indicating weaker OOD performance.}
\label{fig:drop}
\end{figure}

OOD generalization remains a major challenge in AI manipulation localization. 
Even under some shared-backbone multitask settings, models exhibit a substantially larger cross-domain drop in localization F1 than in classification F1, as shown in Fig.~\ref{fig:drop}.
It therefore raises a fundamental question: \textbf{why is manipulation localization substantially more sensitive to domain shifts than classification?}
As discussed in Sec.~2, tighter alignment between real and fake image distributions reduces dataset-specific discrepancies and encourages detectors to learn more transferable forensic evidence. 
However, compared with image-level labels, pixel-level supervision introduces substantially more non-causal factors into the learning objective. 
First, image-level classification only needs to determine whether manipulation traces exist, whereas localization must further associate these traces with precise spatial positions. 
Second, manipulation masks inevitably encode dataset-specific properties, including shape, position, area, and semantic boundaries. 
These properties can become highly predictive spatial shortcuts, causing models to rely on mask-related correlations rather than intrinsic forensic traces. 
Consequently, manipulation localization is more sensitive to domain shifts and exhibits weaker OOD generalization.

\subsection{\dataset for Alignment}

\begin{figure}[t]
\centering
\includegraphics[width=\columnwidth]{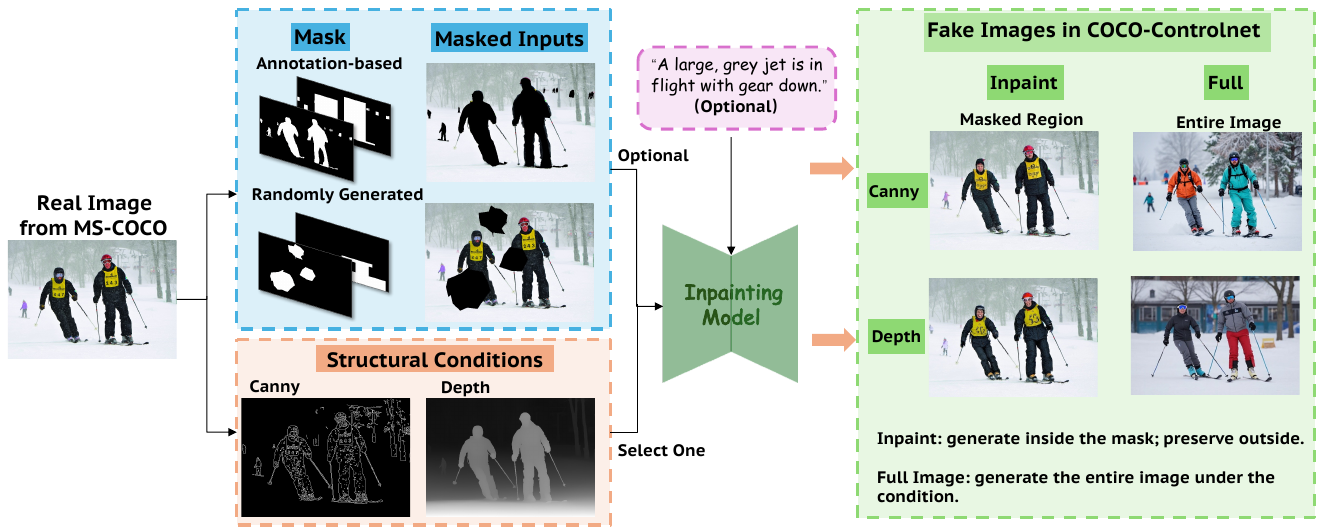}
\caption{
COCO-ControlNet data construction pipeline.
Starting from a real MS-COCO image, the pipeline combines an annotation-based or randomly generated mask, an optional text prompt, and either Canny or depth conditioning to produce an inpainted region or a fully generated image.
}
\label{fig:data}
\end{figure}

The above analysis suggests that manipulation localization would benefit from distribution alignment and may require even tighter alignment. 
However, most existing manipulation datasets perform masked editing on semantic objects in real images, which can introduce noncausal cues such as semantic boundaries. 
In addition, manipulation models generate manipulated regions solely from the text prompt and the unmasked context, often leading to unstable synthesis and visible artifacts. 
To achieve better alignment, we explore ControlNet~\cite{zhang2023adding}, which injects structured geometric conditions from the source image to better preserve scene structure. 
Specifically, we construct \dataset following the data construction pipeline and protocol of COCO-Inpaint~\cite{yan2025cocoinpaint}. 
COCO-Inpaint uses four mask types, including random masks, which partially alleviate shortcut cues associated with semantic boundaries. 
We use FLUX.1 [dev]~\cite{flux1dev2024} as the base generative model: inpainting is constrained by the manipulation mask, while both inpainting and full-image generation use either Canny or depth ControlNet conditioning.
This configuration produces both inpainting and full-image generation results, as shown in Fig.~\ref{fig:data}.
COCO-ControlNet contains $\sim$551K training samples and $\sim$23K validation samples.

\subsection{Is There a Better Alignment Strategy?}

\begin{figure}[t]
\centering
\includegraphics[width=\columnwidth]{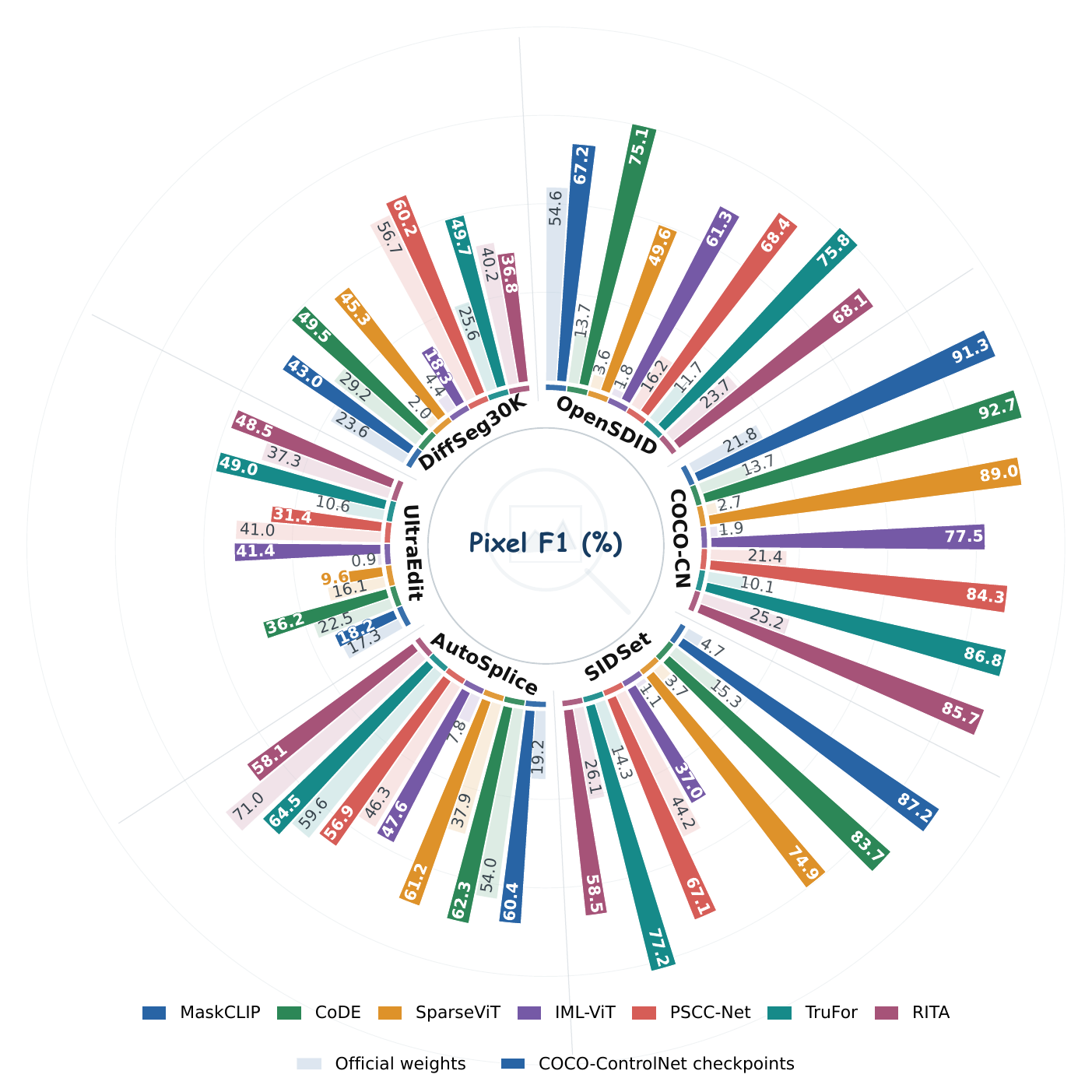}
\caption{Pixel-F1 comparison between official weights and checkpoints retrained on COCO-ControlNet.}
\label{fig:coco-better-pixel-f1}
\end{figure}

As shown in Fig.~\ref{fig:coco-better-pixel-f1}, all seven baselines retrained on COCO-ControlNet achieve higher mean OOD Pixel-F1 than their official checkpoints.
This consistent result supports our hypothesis that distribution-aligned training can improve OOD localization.
This result raises a natural question: \textbf{can tighter alignment yield further gains? }

To explore this question, we propose a new extreme pixel-alignment strategy, \textbf{Mask-VAE}. 
Specifically, we replace the masked region of a real image with the corresponding region from its VAE reconstruction and treat the resulting image as a locally manipulated sample. 
Table~\ref{tab:mask-vae-comparison} shows that COCO-ControlNet training outperforms Mask-VAE training in every reported column for both CoDE and MaskCLIP.
This result suggests that maximizing pixel similarity does not necessarily yield optimal generalization.

\begin{table}[t]
\centering
\small
\setlength{\tabcolsep}{2.5pt}
\begin{tabular}{@{}llccc@{}}
\toprule
Method & Training data & COCO-CN & OOD Avg. & All Avg. \\
\midrule
CoDE & Mask-VAE & 41.0/25.8 & 44.6/29.8 & 44.0/29.1 \\
 & \textbf{COCO-CN} & \textbf{92.7/86.4} & \textbf{61.4/46.5} & \textbf{66.6/53.1} \\
\addlinespace
MaskCLIP & Mask-VAE & 35.6/21.7 & 48.7/33.6 & 46.6/31.6 \\
 & \textbf{COCO-CN} & \textbf{91.3/84.1} & \textbf{55.2/41.7} & \textbf{61.2/48.8} \\
\bottomrule
\end{tabular}
\caption{Mask-VAE versus COCO-ControlNet training under matched COCO sources and masks (F1/IoU $\uparrow$).}
\label{tab:mask-vae-comparison}
\end{table}

The substantially lower performance of Mask-VAE on the COCO-ControlNet validation set suggests that VAE reconstruction artifacts transfer poorly to manipulation artifacts produced by local diffusion-based inpainting.
We attribute this limited transferability to two mismatches in spatial support and artifact formation.
First, VAE reconstruction operates on the entire image and introduces decoder traces across its global spatial support, whereas Mask-VAE merely restricts these globally formed traces to the masked region.
It therefore changes where the reconstruction artifacts appear without adapting how they are formed to the boundary and context of a local manipulation.
Second, genuine local diffusion inpainting performs iterative denoising inside the mask while conditioning on the unmasked pixels and other conditioning signals.
Its artifacts are consequently shaped by the generation process, surrounding context, noise schedule, mask boundary, and conditioning information, none of which is reproduced by a simple VAE encode--decode operation.
Thus, although Mask-VAE provides VAE-related forensic cues, its mismatch with genuine local diffusion-inpainting artifacts limits its effectiveness as dense supervision for manipulation localization.

\subsection{Boundary Adhesion in Segmentation Models} 

\begin{figure}[t]
\centering
\includegraphics[width=\columnwidth]{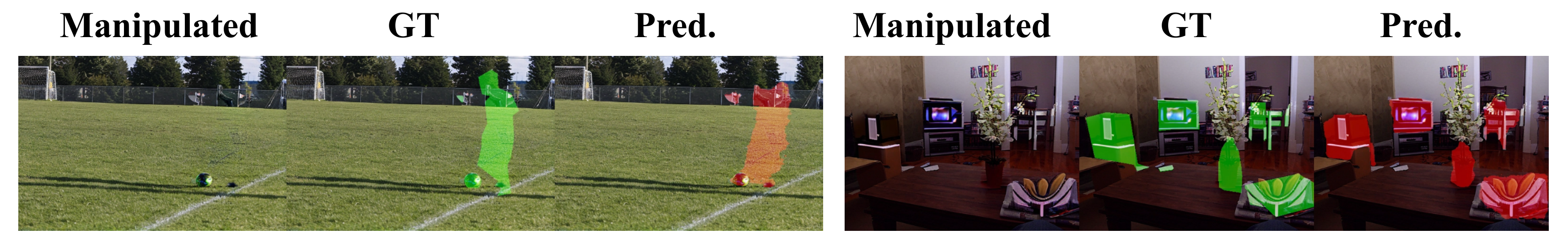}
\caption{
Boundary adhesion in SAM3 directly fine-tuned on COCO-ControlNet.
From left to right: manipulated image, ground-truth mask (green), and predicted mask (red).
The predictions partially overlap the ground-truth regions but expand or shrink to the full contours of salient semantic objects rather than following the true manipulation boundaries.
}
\label{fig:sam3_finetune}
\end{figure}

Most AI manipulation localizers fine-tune pretrained segmentation backbones~\cite{xie2021segformer,kirillov2023segment}.
However, these models are pretrained to recognize semantic objects, whereas manipulation regions do not necessarily follow object boundaries.
When training masks frequently coincide with semantic contours, the localization loss reinforces this inherited prior and encourages the model to treat semantic boundaries as manipulation boundaries.
We call this failure mode \emph{boundary adhesion}.
Conventional fine-tuning relies only on training images, ground-truth masks, and localization losses.
It may therefore associate manipulation boundaries with dataset-specific mask geometry and semantic content, which inherently limits OOD generalization.
As shown in Fig.~\ref{fig:sam3_finetune}, a directly fine-tuned model often identifies the approximate manipulated region, but its prediction expands along the contours of semantic objects such as people, furniture, or road infrastructure instead of following the ground-truth manipulation mask.

Effective localization thus requires preserving the strong spatial representations of segmentation models while suppressing their semantic boundary shortcut. 
To this end, we introduce VAE reconstruction artifacts as a global generation prior. 
Instead of passively fitting the boundary concepts defined by the training data and ground truth masks, the model uses this prior to search for boundaries supported by forensic evidence. 
Since the prior provides global artifact information without directly specifying a spatial mask, it reduces dataset specific overfitting while retaining the localization capacity of the segmentation backbone, thereby improving OOD generalization.

\section{Method}

\begin{figure*}[!t]
\centering
\includegraphics[width=\textwidth]{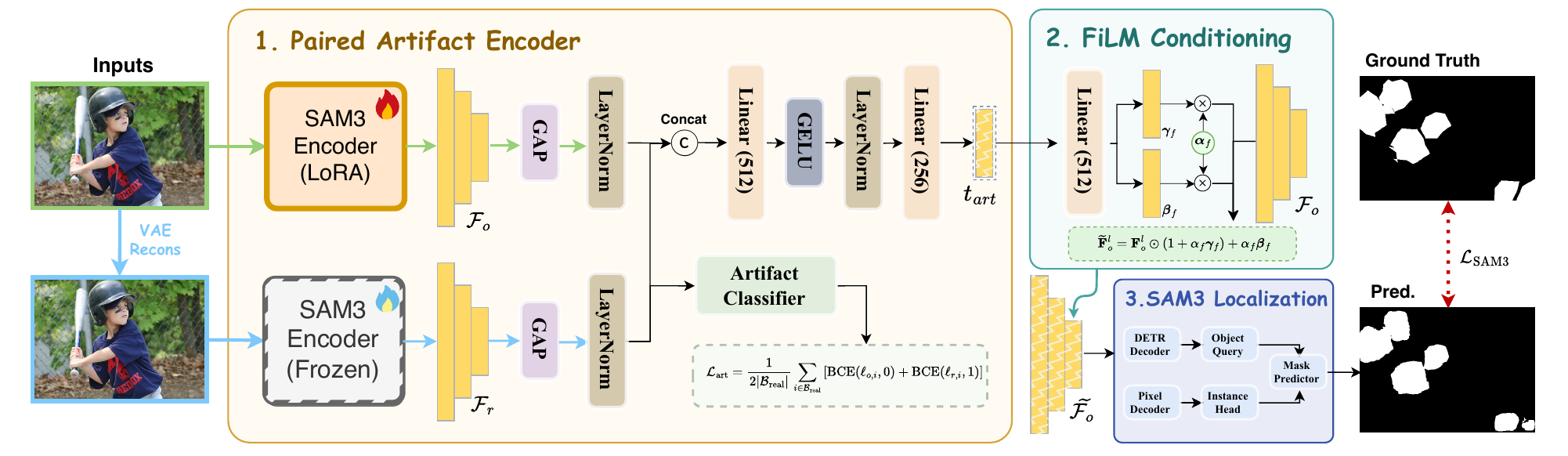}
\caption{
Overview of \method.
Adaptive and frozen SAM3 encoders extract features from the input image and its VAE reconstruction, respectively, and the Paired Artifact Encoder fuses the pooled features into a Global Artifact Prior Token supervised by an auxiliary artifact classifier.
Zero-gated FiLM uses this token to modulate the input-image feature pyramid before the native SAM3 localization pipeline predicts the manipulation mask.
}
\label{fig:method}
\end{figure*}

As discussed above, VAE reconstruction provides image-wide forensic evidence but no spatially aligned manipulation cue, while directly fine-tuned segmentation models can overfit the semantic boundary patterns associated with training masks.
To mitigate this mismatch, we propose \method, a SAM3-based manipulation localizer that encodes the input image and its reconstruction from a frozen VAE into a global artifact token and injects this prior into SAM3's feature pyramid before mask decoding.
As shown in Fig.~\ref{fig:method}, \method comprises three components: the Global Artifact Prior, FPN Mask Decoding, and the Learning Objective, which are detailed in the following subsections.

\subsection{Global Artifact Prior}
\label{sec:global_artifact_prior}

Given a batch of input images $\mathbf{x}\in\mathbb{R}^{B\times3\times H\times W}$, we encode each image and its frozen VAE reconstruction using adaptive and frozen configurations of the same pretrained SAM3 image encoder, respectively.
The two branches produce $L$-level FPN feature pyramids $\mathcal{F}_o$ and $\mathcal{F}_r$, respectively:
\begin{equation}
\begin{aligned}
\mathcal{F}_o&=\Phi_{\theta}(\mathbf{x})=\{\mathbf{F}_o^l\}_{l=1}^{L},\\
\mathcal{F}_r&=\Phi_0\!\left(\mathcal{R}(\mathbf{x})\right)=\{\mathbf{F}_r^l\}_{l=1}^{L}.
\end{aligned}
\label{eq:paired_encoding}
\end{equation}
The 256-dimensional final-layer maps $\mathbf{F}_o^L$ and $\mathbf{F}_r^L$ form the paired input to the Paired Artifact Encoder, while the full observed-image pyramid $\mathcal{F}_o$ is retained for localization.
The reconstructed image is processed by a separate frozen encoder to decouple global artifact-prior extraction from localization-oriented feature learning, preventing localization training from entangling the two representations.

\begin{table*}[!t]
\centering
\small
\setlength{\tabcolsep}{2.5pt}
\renewcommand{\arraystretch}{1.12}
\newcommand{\tablecontinuousrule}{\smash{\rule[-0.85ex]{\arrayrulewidth}{3.45ex}}}
\begin{tabular}{@{}l!{\tablecontinuousrule}cc!{\tablecontinuousrule}cc:cc:cc:cc:cc!{\tablecontinuousrule}cc@{}}
\specialrule{1pt}{0pt}{0pt}
\rule{0pt}{2.6ex} & \multicolumn{2}{c!{\tablecontinuousrule}}{COCO-CN} & \multicolumn{2}{c:}{OpenSDID} & \multicolumn{2}{c:}{SID-Set} & \multicolumn{2}{c:}{AutoSplice} & \multicolumn{2}{c:}{UltraEdit} & \multicolumn{2}{c!{\tablecontinuousrule}}{DiffSeg30K} & \multicolumn{2}{c}{Avg.} \\
Method\rule[-0.8ex]{0pt}{2.4ex} & F1 & IoU & F1 & IoU & F1 & IoU & F1 & IoU & F1 & IoU & F1 & IoU & F1 & IoU \\
\hline
SparseViT~\cite{su2025can} & 89.0 & 80.1 & 49.6 & 33.0 & 74.9 & 59.8 & 61.2 & 44.1 & 9.6 & 5.1 & 45.3 & 29.3 & 54.9 & 41.9 \\
IML-ViT~\cite{ma2023imlvit} & 77.5 & 63.3 & 61.4 & 44.2 & 37.0 & 22.7 & 47.6 & 31.3 & 41.4 & 26.1 & 18.3 & 10.1 & 47.2 & 32.9 \\
PSCC-Net~\cite{liu2022pscc} & 84.3 & 72.9 & 68.4 & 52.0 & 67.1 & 50.5 & 56.9 & 39.8 & 31.4 & 18.7 & 60.2 & 43.1 & 61.4 & 46.2 \\
TruFor~\cite{guillaro2023trufor} & 86.8 & 76.6 & \underline{75.8} & \underline{61.1} & 77.2 & 62.9 & \underline{64.5} & \underline{47.6} & \underline{49.0} & \underline{32.5} & 49.7 & 33.1 & \underline{67.2} & 52.3 \\
CoDE~\cite{Peng2024CoDETIFS} & \underline{92.7} & \underline{86.4} & 75.1 & 60.2 & 83.7 & 71.9 & 62.3 & 45.2 & 36.2 & 22.1 & 49.5 & 32.9 & 66.6 & \underline{53.1} \\
MaskCLIP (Wang et al.~\citeyear{wang2025opensdi}) & 91.3 & 84.1 & 67.3 & 50.7 & \underline{87.2} & \underline{77.3} & 60.4 & 43.2 & 18.2 & 10.0 & 43.0 & 27.4 & 61.2 & 48.8 \\
SIDA~\cite{Huang2025SIDA} & 82.6 & 70.3 & 31.4 & 18.6 & 63.2 & 46.1 & 56.1 & 39.0 & 36.9 & 22.6 & \underline{67.0} & \underline{50.4} & 56.2 & 41.2 \\
RITA~\cite{Zhu2026RITACVPRF} & 85.7 & 75.0 & 68.1 & 51.6 & 58.5 & 41.4 & 58.1 & 40.9 & 48.5 & 32.0 & 36.8 & 22.6 & 59.3 & 43.9 \\
\rule[-0.8ex]{0pt}{2.4ex}\textbf{\method} & \textbf{98.3} & \textbf{96.7} & \textbf{90.5} & \textbf{82.7} & \textbf{92.7} & \textbf{86.3} & \textbf{78.3} & \textbf{64.4} & \textbf{49.8} & \textbf{33.2} & \textbf{69.0} & \textbf{52.7} & \textbf{79.8} & \textbf{69.3} \\
\specialrule{1pt}{0pt}{0pt}
\end{tabular}%
\caption{Pixel-level localization after training on COCO-ControlNet, with COCO-ControlNet as the in-domain test set and all other test sets as OOD benchmarks.
Best and second-best results are bolded and underlined, respectively.}
\label{tab:sota}
\end{table*}

\paragraph{Paired Artifact Encoder.}
\label{sec:paired_artifact_encoder}

The Paired Artifact Encoder applies global average pooling (GAP) and branch-specific LayerNorm to the paired final-layer feature maps $\mathbf{F}_o^L$ and $\mathbf{F}_r^L$ in Eq.~\ref{eq:paired_encoding}, yielding two 256-dimensional descriptors:
\begin{equation}
\mathbf{h}_{o,r}=\operatorname{LN}_{o,r}(\operatorname{GAP}(\mathbf{F}_{o,r}^L)).
\label{eq:artifact_descriptors}
\end{equation}
It then concatenates the descriptors and applies two linear layers with GELU and LayerNorm in between to produce the 256-dimensional Global Artifact Prior Token $\mathbf{t}_{\mathrm{art}}$:
\begin{equation}
\mathbf{t}_{\mathrm{art}}
=\operatorname{LN}\!\left(
\operatorname{GELU}\!\left(
[\mathbf{h}_o;\mathbf{h}_r]\mathbf{W}_1+\mathbf{b}_1
\right)\right)\mathbf{W}_2+\mathbf{b}_2,
\label{eq:gap_token}
\end{equation}
where $\mathbf{W}_1\in\mathbb{R}^{512\times512}$ and $\mathbf{W}_2\in\mathbb{R}^{512\times256}$ are learnable weights, $\mathbf{b}_1$ and $\mathbf{b}_2$ are their biases, and $[\,;\,]$ denotes feature concatenation.
The first linear layer lets observed-image and reconstruction information interact in a 512-dimensional joint space, while the second compresses the fused representation into a 256-dimensional token.
Unlike fixed feature differencing, this learned fusion preserves both view-specific states while modeling their asymmetric relationship.
We apply GAP before fusion to remove spatial coordinates and represent reconstruction-related artifact evidence, rather than a localization proposal, in $\mathbf{t}_{\mathrm{art}}$.
The resulting global prior guides the SAM3 Mask Decoder toward manipulation boundaries consistent with the artifact evidence, rather than merely fitting those defined by the training data and ground-truth masks.
Meanwhile, $\mathbf{h}_o$ and $\mathbf{h}_r$ are passed to the Artifact Classifier in Section~\ref{sec:learning_objective} to explicitly constrain the generation attributes of the paired features.
\subsection{FPN Mask Decoding}

The purpose of prior injection is to prevent mask supervision from being the decoder's only guide to forensic evidence.
Under direct fine tuning, the FPN can become strongly adapted to the spatial layouts and semantic boundaries present in the training masks.
We therefore use zero-gated FiLM to inject the global artifact prior into the multiscale FPN.

\paragraph{FiLM Conditioning.}

Starting from the Global Artifact Prior Token $\mathbf{t}_{\mathrm{art}}$ and multiscale pyramid $\mathcal{F}_o$ in Section~\ref{sec:global_artifact_prior}, we apply feature-wise linear modulation (FiLM)~\cite{perez2018film} channel-wise to each FPN level:
\begin{equation}
\widetilde{\mathbf{F}}_o^{l}
=\mathbf{F}_o^{l}\odot
\left(1+\alpha_f\boldsymbol{\gamma}_f\right)
+\alpha_f\boldsymbol{\beta}_f,
\label{eq:fpn_film}
\end{equation}
We project $\mathbf{t}_{\mathrm{art}}$ into channel-wise scale and bias vectors using learnable parameters $\mathbf{W}_f^{\mathrm{FPN}}$ and $\mathbf{b}_f^{\mathrm{FPN}}$:
\begin{equation}
[\boldsymbol{\gamma}_f;\boldsymbol{\beta}_f]
=\mathbf{t}_{\mathrm{art}}\mathbf{W}_f^{\mathrm{FPN}}
+\mathbf{b}_f^{\mathrm{FPN}}.
\label{eq:fpn_film_parameters}
\end{equation}
Both vectors are reshaped and broadcast over the spatial dimensions of each FPN level.
The shared learnable scalar gate $\alpha_f$ is initialized to zero, making the modulation an identity transformation at the beginning of training.
As training proceeds, $\alpha_f$ controls how strongly the global prior modifies the pretrained FPN features.
Conditioning the multiscale features on $\mathbf{t}_{\mathrm{art}}$ supplies an image-specific generation prior before dense decoding: ground-truth masks still supervise where the manipulation lies, but they no longer solely determine what evidence the spatial representation should emphasize.
Because the modulation is broadcast over spatial locations, it supplies forensic context without inserting a reconstruction-derived region or boundary.

SAM3 follows a DETR-style architecture, and we retain its native mask-decoding pipeline without modifying the decoder structure.
Injecting the global artifact prior at the entrance of the Mask Decoder, before the Pixel Decoder, allows zero-initialized FiLM to modulate all multiscale feature maps channel-wise and enables subsequent decoding to propagate and amplify the prior.
Both the query and pixel branches consume the same conditioned feature pyramid: across the DETR Decoder layers, the object queries repeatedly attend to these visual features through cross-attention, while the Pixel Decoder transforms them into dense pixel embeddings.
The artifact prior therefore reaches both the refined queries and dense pixel embeddings before the Query Mask Predictor combines them to produce the final masks $\widehat{\mathbf{M}}$.

\subsection{Learning Objective}
\label{sec:learning_objective}

\paragraph{Localization Objective.}
To preserve SAM3's localization capability, we adopt its official fine-tuning configuration and denote the localization objective by $\mathcal{L}_{\mathrm{SAM3}}$.

\paragraph{Artifact Classifier.}
Localization supervision alone does not ensure that the paired representation captures generation evidence rather than image content.
We therefore attach a shared Artifact Classifier $c(\cdot)$ to the pooled descriptors inside the Paired Artifact Encoder:
\begin{equation}
\ell_o=c(\mathbf{h}_o),\qquad
\ell_r=c(\mathbf{h}_r),
\label{eq:artifact_logits}
\end{equation}
where $\mathbf{h}_o$ and $\mathbf{h}_r$ are the pooled original and reconstruction features, respectively.
Under the real-only setting, let $\mathcal{B}_{\mathrm{real}}$ denote the authentic samples in a batch.
The classifier treats each authentic original as real and its VAE reconstruction as synthetic:
\begin{equation}
\mathcal{L}_{\mathrm{art}}
=\frac{1}{2|\mathcal{B}_{\mathrm{real}}|}
\sum_{i\in\mathcal{B}_{\mathrm{real}}}
\left[
\operatorname{BCE}(\ell_{o,i},0)
+\operatorname{BCE}(\ell_{r,i},1)
\right].
\label{eq:artifact_loss}
\end{equation}
The complete training objective is
\begin{equation}
\mathcal{L}
=\mathcal{L}_{\mathrm{SAM3}}
+\lambda_{\mathrm{art}}\mathcal{L}_{\mathrm{art}}.
\label{eq:full_objective}
\end{equation}
The Artifact Classifier anchors the paired feature space to the real-versus-reconstructed generation axis, making $\mathbf{t}_{\mathrm{art}}$ a more explicit forensic prior for the downstream decoder.
Because it acts only on globally pooled descriptors, it introduces no mask, boundary, or other spatial supervision.
No type-classification, semantic, or auxiliary boundary loss is introduced.

\section{Experiments}

\begin{figure*}[!t]
\centering
\includegraphics[width=\textwidth]{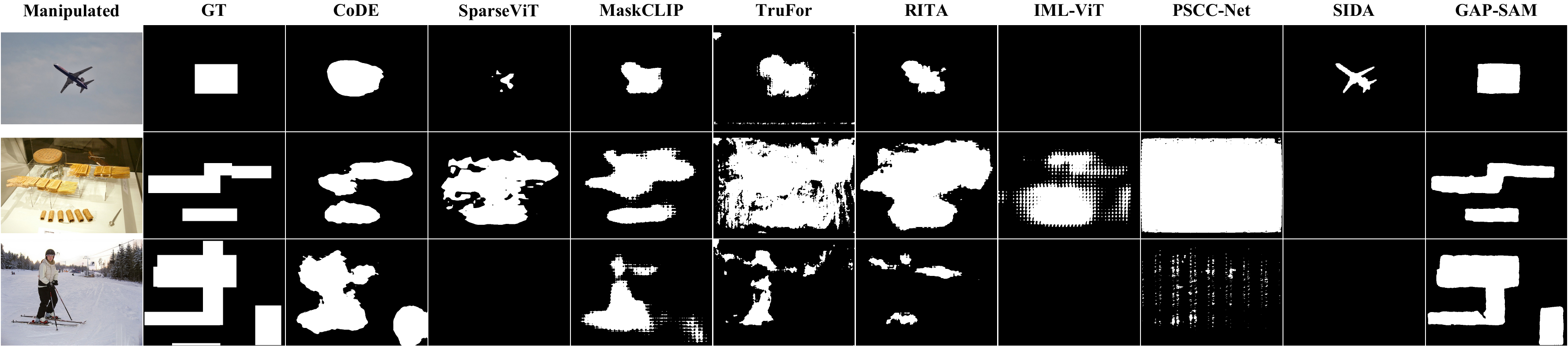}
\caption{Qualitative comparison: \method reduces semantic boundary adhesion and localizes arbitrary manipulation regions.}
\label{fig:qualitative}
\end{figure*}

\subsection{Implementations.}

\paragraph{Implementation details.}
All methods are trained on the same \dataset training split.
For each re-trained baseline, we retain its default training hyperparameters and change only the training data; validation loss is evaluated every 0.1 epoch, with early stopping using a patience of five evaluations.
The \dataset and Mask-VAE variants use the same COCO source images and manipulation masks.
The VAE used for all reconstructions is taken from Stable Diffusion 2.1 (SD2.1)~\cite{rombach2022ldm}.
For \method, the SAM3 LoRA adapters use rank 8, scaling factor 16, and zero dropout, while the other training settings follow the official SAM3 recipe~\cite{carion2025sam3}.
We train with AdamW for 3 epochs using an effective batch size of 32, a learning rate of $2\times10^{-4}$, weight decay 0.05, and $\lambda_{\mathrm{art}}=1.0$.

\paragraph{OOD benchmarks.}
We evaluate on six test sets: the held-out \dataset split is the in-domain benchmark, while OpenSDID~\cite{wang2025opensdi}, SID-Set~\cite{Huang2025SIDA}, AutoSplice~\cite{Jia2023AutoSpliceCVPRW}, UltraEdit~\cite{zhao2024ultraedit}, and DiffSeg30K~\cite{Ci2025DiffSeg30kArxiv} constitute five OOD benchmarks.
We report Pixel-F1 and intersection over union (IoU) at a fixed threshold of 0.5, and compute the OOD result as the unweighted average over the five OOD benchmarks.

\paragraph{Baselines.}
We compare with conventional forensic localizers, including PSCC-Net~\cite{liu2022pscc}, TruFor~\cite{guillaro2023trufor}, and CoDE~\cite{Peng2024CoDETIFS}; Transformer-based localizers IML-ViT~\cite{ma2023imlvit} and SparseViT~\cite{su2025can}; the pretrained vision-model fusion method MaskCLIP~\cite{wang2025opensdi}; the process-aware autoregressive localizer RITA~\cite{Zhu2026RITACVPRF}; and the large-multimodal-model-based SIDA~\cite{Huang2025SIDA}.

\subsection{Comparison with the State of the Art}

Table~\ref{tab:sota} reports pixel-level localization after training all methods on COCO-ControlNet.
Existing methods vary substantially across datasets: TruFor is relatively strong on UltraEdit, while CoDE obtains the second-highest overall IoU.
Across all six test sets, \method achieves the highest Pixel-F1 and IoU, improving the six-set average over the strongest prior result for each metric by 12.6 F1 and 16.2 IoU points, respectively.
Qualitative comparisons in Figure~\ref{fig:qualitative} show that for edits not aligned with semantic boundaries, \method reduces boundary adhesion and emphasizes boundaries supported by forensic evidence.
GAP conditioning preserves precise SAM3 contours when they agree with such evidence, while avoiding default selection of the entire semantic instance and recovering low-texture background edits and masks spanning multiple objects.

\subsection{Ablation Studies}

\paragraph{Incremental module additions.}
Table~\ref{tab:ablation-components} incrementally adds the proposed modules to SAM3 LoRA.
For the original-only prior, both encoder branches process the observed image rather than using its VAE reconstruction in the frozen branch, separating the effect of introducing the prior branch from the additional contribution of VAE reconstruction evidence.
OOD performance improves consistently as the modules are added, validating the effectiveness of each component.
Conditioning on VAE reconstruction evidence provides the key gain, while the Artifact Loss further refines this signal to represent generation artifacts more accurately.

\begin{table}[t]
\centering
\small
\setlength{\tabcolsep}{2.5pt}
\begin{tabular}{@{}lccc@{}}
\toprule
Setting (F1/IoU $\uparrow$) & COCO-CN & OOD Avg. & All Avg. \\
\midrule
Baseline & 95.4/91.1 & 68.5/54.4 & 73.0/60.6 \\
+ Original-only prior & 95.7/92.1 & 69.4/55.1 & 73.8/61.3 \\
+ Reconstruction-paired prior & 97.3/94.8 & 73.5/59.7 & 77.5/65.6 \\
\textbf{+ Artifact supervision (Full)} & \textbf{98.3/96.7} & \textbf{76.1/63.9} & \textbf{79.8/69.3} \\
\bottomrule
\end{tabular}
\caption{
Incremental module ablation.
}
\label{tab:ablation-components}
\end{table}

\paragraph{Conditioning operator.}
Table~\ref{tab:ablation-conditioning} compares the operators used to apply the global token.
FiLM achieves the highest COCO-CN, OOD, and six-set averages and is used in the reference design.

\begin{table}[t]
\centering
\small
\setlength{\tabcolsep}{2.5pt}
\begin{tabular}{@{}lccc@{}}
\toprule
Setting (F1/IoU $\uparrow$) & COCO-CN & OOD Avg. & All Avg. \\
\midrule
Gated concatenation & 96.6/93.7 & 75.0/62.4 & 78.6/67.6 \\
Residual addition & 95.4/94.2 & 74.7/61.4 & 78.2/66.9 \\
\textbf{FiLM} & \textbf{98.3/96.7} & \textbf{76.1/63.9} & \textbf{79.8/69.3} \\
\bottomrule
\end{tabular}
\caption{
Conditioning operator ablation.
}
\label{tab:ablation-conditioning}
\end{table}

\paragraph{Prior injection site.}
Table~\ref{tab:ablation-injection} compares prior-injection sites while keeping the native query-mask output fixed.
Conditioning the FPN before the Pixel Decoder provides the best OOD generalization and is therefore used by default.

\begin{table}[t]
\centering
\small
\setlength{\tabcolsep}{2.5pt}
\begin{tabular}{@{}lccc@{}}
\toprule
Injection site (F1/IoU $\uparrow$) & COCO-CN & OOD Avg. & All Avg. \\
\midrule
Object queries & 96.6/93.4 & 72.2/60.6 & 76.3/66.1 \\
Post-decoder features & 97.4/95.1 & 74.4/62.0 & 78.2/67.5 \\
Instance embedding & \textbf{98.5/97.1} & 74.7/62.8 & 78.7/68.5 \\
\textbf{Pre-decoder FPN} & 98.3/96.7 & \textbf{76.1/63.9} & \textbf{79.8/69.3} \\
\bottomrule
\end{tabular}
\caption{Injection-site ablation with query-mask output fixed; the last row is the default \method configuration.}
\label{tab:ablation-injection}
\end{table}

\paragraph{Encoder and Prior Representation.}
Table~\ref{tab:ablation-encoder-prior} jointly examines Reconstruction Encoder adaptation and artifact-prior representations.
The frozen encoder and global pooling offer the strongest or comparable OOD generalization while keeping the design simple and the reference features stable, and are therefore used by default.

\begin{table}[t]
\centering
\small
\setlength{\tabcolsep}{2.5pt}
\begin{tabular}{@{}lccc@{}}
\toprule
Setting (F1/IoU $\uparrow$) & COCO-CN & OOD Avg. & All Avg. \\
\midrule
\multicolumn{4}{@{}l}{\textbf{Reconstruction Encoder}} \\
Shared LoRA & \textbf{98.7/97.5} & 73.0/61.3 & 77.3/67.3 \\
\textbf{Frozen} & 98.3/96.7 & \textbf{76.1/63.9} & \textbf{79.8/69.3} \\
\addlinespace
\multicolumn{4}{@{}l}{\textbf{Artifact-Prior Representation}} \\
Spatial delta & \textbf{99.0/98.1} & 74.6/62.9 & 78.7/68.8 \\
Spatial cross-attn. & 97.6/95.4 & 74.4/60.9 & 78.3/66.6 \\
Multi-level pooling & 98.5/97.0 & \textbf{76.3}/63.7 & \textbf{80.0/69.3} \\
\textbf{Global pooling} & 98.3/96.7 & 76.1/\textbf{63.9} & 79.8/\textbf{69.3} \\
\bottomrule
\end{tabular}
\caption{Reconstruction encoder/GAP encoding ablation.}
\label{tab:ablation-encoder-prior}
\end{table}

\subsection{Robustness}

We evaluate robustness to JPEG compression, Gaussian blur, and resizing on COCO-ControlNet.
\method remains robust across all three distortions and outperforms all baselines.

\begin{figure}[!t]
    \centering
    \begin{minipage}{\columnwidth}
        \centering
        \includegraphics[width=0.99\textwidth]{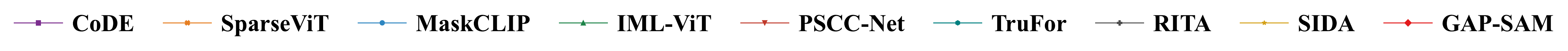}
        \begin{minipage}[b]{0.32\textwidth}
            \includegraphics[width=\textwidth]{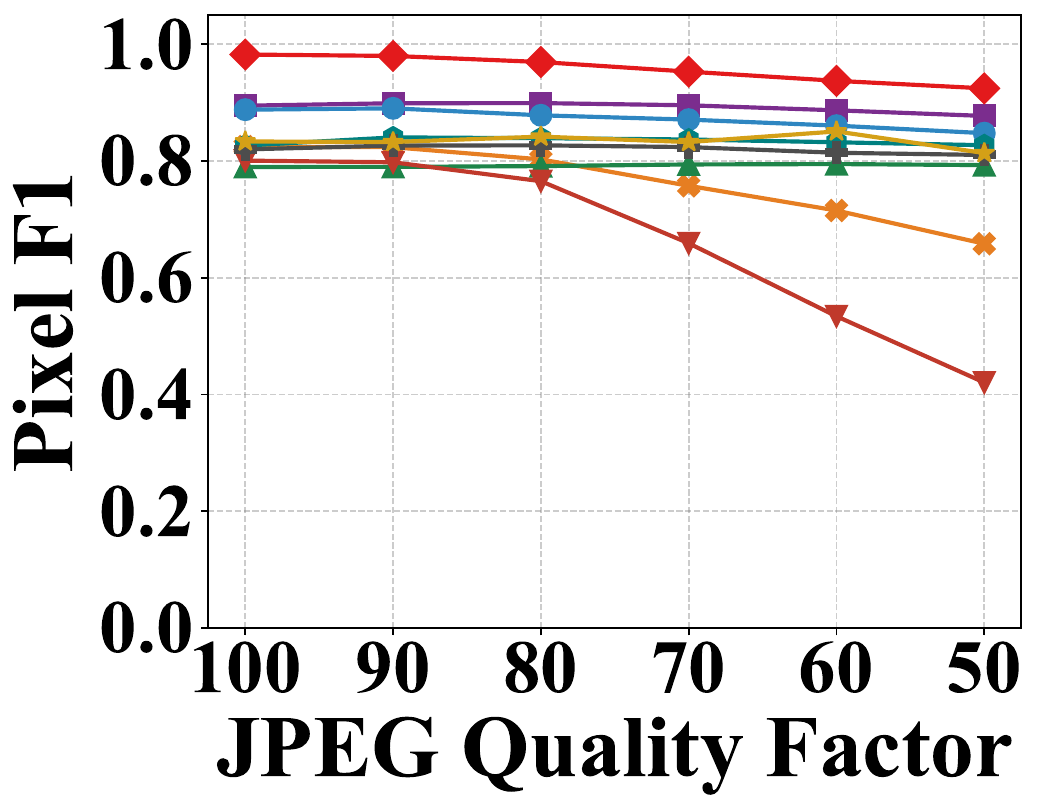}
        \end{minipage}%
        \hfill%
        \begin{minipage}[b]{0.32\textwidth}
            \includegraphics[width=\textwidth]{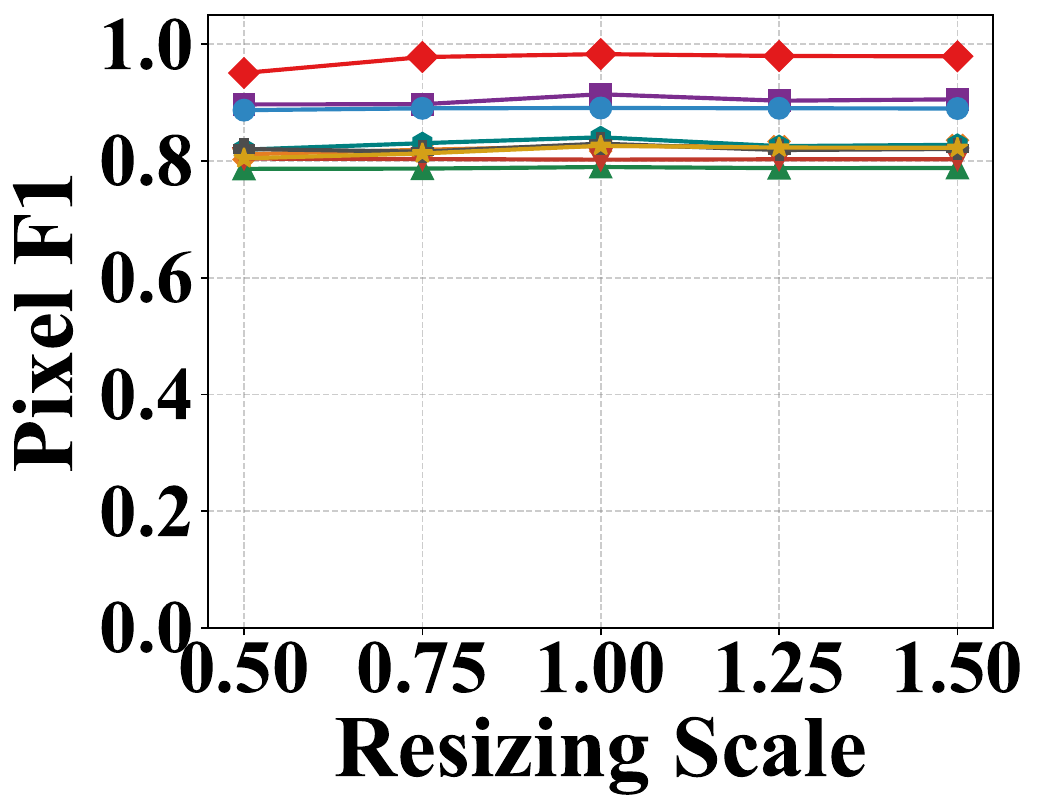}
        \end{minipage}%
        \hfill%
        \begin{minipage}[b]{0.32\textwidth}
            \includegraphics[width=\textwidth]{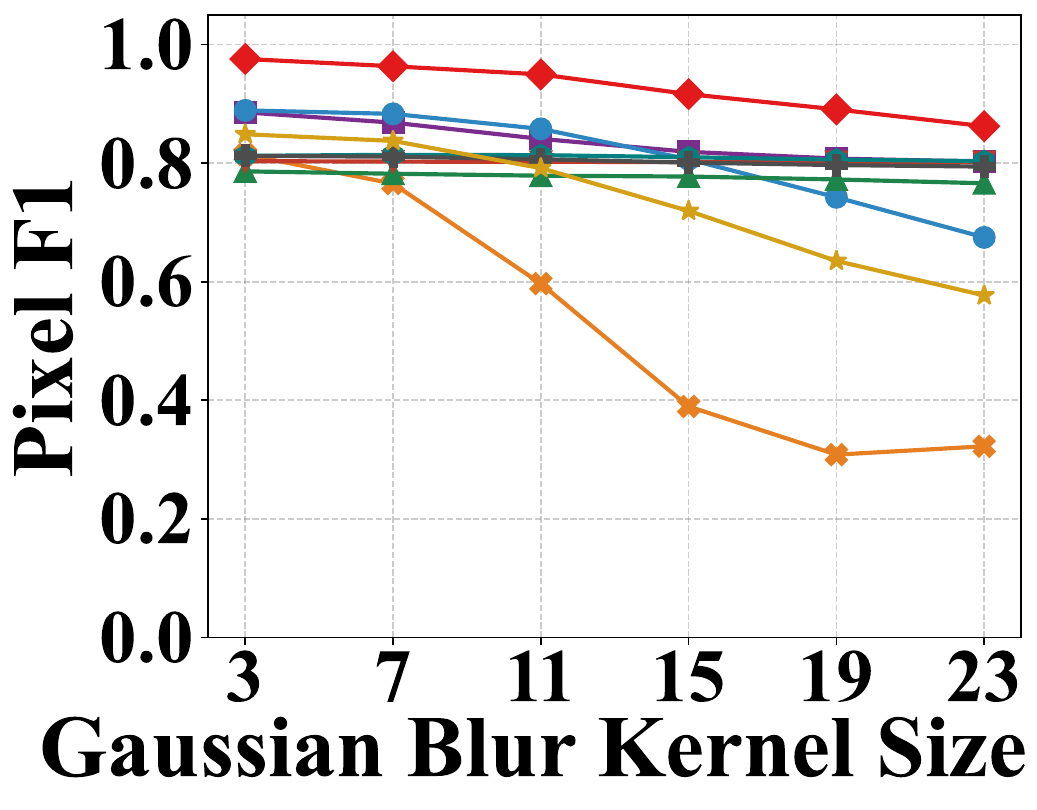}
        \end{minipage}
    \end{minipage}
    \caption{Robustness analysis on COCO-ControlNet under JPEG compression, resizing, and Gaussian blur.}
\label{fig:robustness}
\end{figure}

\section{Conclusion}

We studied why OOD generalization deteriorates when AI detection moves from image labels to pixel masks.
The Mask-VAE comparison further suggests that alignment alone is insufficient when the artifact formation process and its spatial support differ from the localization target.
\dataset uses structural conditioning to reduce semantic shortcuts while preserving diffusion-based inpainting artifacts.
\method then repurposes the globally transferable evidence of a frozen VAE reconstruction as a compact prior for SAM3's feature pyramid.
Zero-initialized FiLM conditioning and artifact classification steer the decoder away from semantic boundary adhesion without supplying a spatial shortcut.
The resulting framework offers a simple bridge between robust whole-image detection and precise manipulation localization.

\bibliography{aaai2027}

\end{document}